%% file: main.tex
\def\year{2022}\relax
\documentclass[letterpaper]{article} 
\usepackage{aaai22}  
\usepackage{times}  
\usepackage{helvet}  
\usepackage{courier}  
\usepackage[hyphens]{url}  
\usepackage{graphicx} 
\usepackage{natbib}  
\usepackage{caption} 
\DeclareCaptionStyle{ruled}{labelfont=normalfont,labelsep=colon,strut=off} 
\usepackage{booktabs}       
\usepackage{amsfonts}       
\usepackage{nicefrac}       
\usepackage{microtype}      

\usepackage{xcolor,colortbl}
\colorlet{newColor}{green!5!orange!95!}

\usepackage{wrapfig,lipsum}
\usepackage{enumitem}
\usepackage{multirow}
\usepackage{hhline}
\usepackage{subcaption}
\usepackage{cancel}
\usepackage{pifont}

\usepackage{makecell}

\usepackage{algorithm}
\usepackage{algorithmic}
\usepackage{amssymb}
\usepackage{amsmath}

\usepackage{textcomp}
\usepackage{siunitx}

\newcommand{\rulesep}{\unskip\ \vrule\ }
\newcommand{\blambda}{{\boldsymbol{\lambda}}}

\newcommand{\btheta}{{\boldsymbol{\theta}}}

\usepackage{newfloat}
\usepackage{listings}
\floatstyle{ruled}
\newfloat{listing}{tb}{lst}{}
\floatname{listing}{Listing}
\title{Shifting Transformation Learning for Out-of-Distribution Detection} 
\author{
    Sina Mohseni, 
    Arash Vahdat, 
    Jay Yadawa
}
\affiliations{

    NVIDIA\\
    USA\\
    \{smohseni, avahdat, jyadawa\}@nvidia.com
}

\usepackage{bibentry}

\begin{document}

\maketitle

\input{sections/0-abstract.tex}

\input{sections/1-intro}

\input{sections/2-background.tex}

\input{sections/3-Method.tex}


\input{sections/5.1.performance}

\input{sections/5.2.robustness}
\input{sections/7-conclusion}
\bibstyle{aaai22}
\bibliography{biblo}

\setcounter{secnumdepth}{1} 

\input{sections/9-appendix}

\end{document}

%% file: sections/0-abstract.tex
\begin{abstract}
Detecting out-of-distribution (OOD) samples plays a key role in open-world and safety-critical applications such as autonomous systems and healthcare. Recently, self-supervised representation learning techniques (via contrastive learning and pretext learning) have shown effective in improving OOD detection. However, one major issue with such approaches is the choice of shifting transformations and pretext tasks which depends on the in-domain distribution.
In this paper, we propose a simple framework that leverages a \textit{shifting transformation learning} setting for learning multiple shifted representations of the training set for improved OOD detection. To address the problem of selecting optimal shifting transformation and pretext tasks, we propose a simple mechanism for automatically selecting the transformations and modulating their effect on representation learning without requiring any OOD training samples. In extensive experiments, we show that our simple framework outperforms state-of-the-art OOD detection models on several image datasets. We also characterize the criteria for a desirable OOD detector for real-world applications and demonstrate the efficacy of our proposed technique against state-of-the-art OOD detection techniques. 
\end{abstract}

%% file: sections/1-intro.tex
\section{Introduction}
\label{sec:intro}

Despite advances in representation learning and their generalization to unseen samples, learning algorithms are bounded to perform well on source distribution and vulnerable to out-of-distribution (OOD) or outlier samples.
For example, it has been shown that the piece-wise linear decision boundaries in deep neural network (DNN) with ReLU activation are prune to OOD samples as they can assign arbitrary high confidence values to samples away from the training distribution \cite{hein2019relu}. 
Recent work on machine learning trustworthiness and safety have shown that
OOD detection plays a key role in open-world and safety-critical applications such as autonomous systems \cite{mohseni2019practical} and healthcare \cite{ren2019likelihood}. 
However, OOD detection in high dimensional domains like image data is a challenging task and often requires great computational resource~\cite{gal2016dropout}. 

The recent surge in self-supervised learning techniques shows that learning pretext tasks can result in better semantic understanding of data by learning invariant representations~\cite{dosovitskiy2014discriminative} and can increase model performance in different setups \cite{gidaris2018unsupervised}.
Self-supervised learning has also been shown effective in OOD detection. For example, \citet{golan2018deep} and \citet{Hendrycks2019sv} show that simple geometric transformations improve OOD detection performance, and \citet{tack2020csi} leverage shifting data transformations and contrastive learning for OOD detection. However, these works manually design the transformations and pretext tasks.

Inspired by the recent works, we study the impact of representation learning on OOD detection when training a model on artificially transformed datasets. We observe that training on a diverse set of dataset transformations jointly, termed as \textit{shifting transformation learning} here, further improves the model's ability to distinguish in-domain samples from outliers.
However, we also empirically observe that the choice of effective data transformations for OOD detection depends on the in-domain training set. In other words, the set of transformations effective for one in-domain dataset may no longer be effective for another dataset.

To address this problem, we make the following contributions in this paper:
\textbf{(i)} We propose a simple framework for transformation learning from multiple shifted views of the in-domain training set in both self-supervised and fully-supervised settings (when data labels are available) for OOD detection.
\textbf{(ii)} We propose a framework that selects effective transformation and modulates their impact on representation learning. We demonstrate that the optimally selected transformations result in better representation for both main classification and OOD detection compared to data augmentation-based approaches.
\textbf{(iii)} We propose an ensemble score for OOD detection that leverages multiple transformations trained with a shared encoder. In particular, our technique achieves new state-of-the-art results in OOD detection on multi-class classification by improving averaged area under the receiver operating characteristics (AUROC) +1.3\% for CIFAR-10, +4.37\% for CIFAR-100, and +1.02\% for ImageNet-30 datasets. 
\textbf{(iv)} To the best of our knowledge, this paper is the first to introduce criteria for ideal OOD detection and to analyze a diverse range of techniques along with these criteria. Albeit the simplicity, we show that our proposed approach outperforms the state-of-the-art techniques on robustness and generalization criteria.

%% file: sections/2-background.tex
\section{Related Work}
\label{sec:background}
Here, we review OOD detection methods related to this work:

\paragraph{Distance-based Detection:} 
Distance-based methods use different distance measures between the unknown test sample and source training set in the representation space. 
These techniques involve preprocessing or test-time sampling of the source domain distribution to measure their averaged distance to the novel input sample.  
The popular distance measures include Mahalanobis distance \cite{lee2018simple,sehwag2021ssd}, cosine similarity \cite{techapanurak2020hyperparameter,tack2020csi} and others semantic similarity metrics \cite{rafiee2020unsupervised}.
These techniques usually work well with unlabeled data in unsupervised and one-class classification setups. 
For example, \citet{ruff2018deep} present a deep learning one-class classification approach to minimize the representation hypersphere for source distribution and calculate the detection score as the distance of the outlier sample to the center of the hypersphere. 
Recently, \citet{mukhoti2021deterministic} proposed using distance measures for model features to better disentangle model uncertainty from dataset uncertainty. 
Distance-based methods can benefit from ensemble measurements over input augmentations \cite{tack2020csi} or transformations \cite{bergman2020classification}, network layers \cite{lee2018simple,sastry2019detecting}, or source domain sub-distributions \cite{oberdiek2020detection} to improve detection results.
For instance, \citet{tack2020csi} present a detection score based on combining representation norm with cosine similarity between the outlier samples and their nearest training samples for one-class classification problem. 
They also show that OOD detection can be improved with ensembling over random augmentations, which carries a higher computational cost.

\paragraph{Classification-based Detection:}
These OOD detection techniques avoid costly distance-based and uncertainty estimation techniques (e.g., \citet{gal2016dropout}) by seeking effective representation learning to encode normality together with the main classification task. 
Various detection scores have been proposed including maximum softmax probability \cite{hendrycks2016baseline}, maximum logit scores \cite{hendrycks2019benchmark}, prediction entropy \cite{mohseni2020ood}, and KL-divergence score \cite{Hendrycks2019sv}. 
To improve the detection performance, \cite{lee2017training,hsu2020generalized} proposed a combination of temperature scaling and adversarial perturbation of input samples to calibrate the model to increase the gap between softmax confidence for the inlier and outlier samples. 
Another line of research proposed using auxiliary unlabeled and disjoint OOD training set to improve OOD detection for efficient OOD detection without architectural changes \cite{hendrycks2018deep,mohseni2020ood}. 

Recent work on self-supervised learning shows that adopting pretext tasks results in learning more invariant representations and better semantic understanding of data \cite{dosovitskiy2014discriminative} and which significantly improves OOD detection \cite{golan2018deep}. 
Hendrycks et al. \cite{Hendrycks2019sv} extended self-supervised techniques with a combination of geometric transformation prediction tasks. 
Self-supervised contrastive training \cite{chen2020simple} is also shown to be effective to leverage from multiple random transformations to learn in-domain invariances, resulting in better OOD detection \cite{winkens2020contrastive,sehwag2021ssd,tack2020csi}.

%% file: sections/3-Method.tex
\input{figures/tsne-vis}

\section{Method}
\label{sec:method}


In this paper, we propose shifting transformation learning for explicit and efficient training of in-domain representation for improved OOD detection.  
Intuitively, we simultaneously train a base encoder on \textit{multiple shifting transformations} of the training data using auxiliary self-supervised objectives for unlabeled data and fully-supervised objectives for labeled data.
To illustrate the impact of our approach on OOD detection, Figure~\ref{fig:tsne} shows t-SNE visualization~\cite{van2008visualizing} of the CIFAR-10 examples obtained from ResNet-18~\cite{he2016deep} trained with the cross-entropy loss (left) compared to our multitask transformation learning (right) with Places365 examples as the OOD test set.
The visualization intuitively shows how training with multiple shifted in-domain samples improves the separation between OOD and in-domain samples without the need for any OOD training set. 

In this section, we first present our shifting transformation learning framework. Then, we present our model for selecting optimal transformation for a given in-domain set. Finally, we present our detection score.

\input{figures/trans-set-and-weights}

\subsection{Shifting Transformation Learning} \label{sec:losses}
Our transformation learning technique trains a multi-tasked network using self-supervised and fully-supervised training objectives. We consider a set of geometric (translation, rotation) and non-geometric (blurring, sharpening, color jittering, Gaussian noise, cutout) shifting transformations and we train the network with dedicated loss functions for each transformation.  
For the self-supervised transformation learning, given an unlabeled training set of  $\mathcal{S} = \{(x_i)\}_{i=1}^{M}$, we denote the set of domain invariant transformations $T_{n}$ by $\mathcal{T} = \{T_{n}\}_{n=1}^N$. 
We generate a self-labeled training set $\mathcal{S}_{T_{n}} = \{(T_{n}(x_{i}), \hat{y}_{i})\}_{i=1}^M$ for each self-supervised transformation $T_{n}$ where $\hat{y}_{i}$ are the transformation labels. 
For example, we consider the image rotation task with four levels of \{0$^{\circ}$, 90$^{\circ}$, 180$^{\circ}$, 270$^{\circ}$\} self-labeled rotations and $\hat{y}_{i} \in \{0, 1, 2, 3\}$ in this case. 
The self-supervised loss $\mathcal{L}_{\text{ssl}}$ is the weighted average of loss across all transformations in $\mathcal{T}$:
\begin{equation} \label{eq:eq_ssl}
\mathcal{L}_{\text{ssl}}(\blambda, \btheta) = \frac{1}{N} \sum_{n=1}^{N} \lambda_{n} \sum_{(T_{n}(x_{i}), \hat{y}_{i}) \in S_{T_{n}}} \ell (f^{(n)}_\btheta(T_n({x_{i})}),\hat{y_i}),     
\end{equation}
where $f^{(n)}_\btheta$ is a classification network with parameters $\btheta$ for the $n^{th}$ task, $\blambda = \{\lambda_n\}_{n=1}^N$ are transformation weights, and $l$ is the multi-class cross-entropy loss.
When labels are available, given the labeled training set of  $\mathcal{S} = \{(x_i, y_i)\}_{i=1}^{M}$, we generate transformed copies of the original training sets $\mathcal{S}_{T_{n}}' = \{(T_{n}(x_{i}), y_i)\}_{i=1}^M$ where training samples retain their original class labels. 
The supervised loss $\mathcal{L}_{\text{sup}}$ is defined by: 
\begin{equation} \label{eq:eq_sup}
\mathcal{L}_{\text{sup}}(\blambda', \btheta) = \frac{1}{N} \sum_{n=1}^{N} \lambda_{n}' \sum_{(T_{n}(x_{i}), y_i) \in S_{T_{n}}'} \ell (f'^{(n)}_\btheta(T_n({x_{i})}),y_i),    
\end{equation}
%
%
%
which measures the classification loss for transformed copies of the data with $\blambda' = \{\lambda'_n\}_{n=1}^N$ as transformation coefficients in $\mathcal{L}_{\text{sup}}$.  
In labeled setup, we combine $\mathcal{L}_{\text{ssl}}$ and $\mathcal{L}_{\text{sup}}$ with the main supervised learning loss $\mathcal{L}_{\text{main}}$ (e.g., the cross-entropy loss for classifying the in-domain training set):
%
\begin{equation} \label{eq:eq_total}
\mathcal{L}_{\text{total}}(\blambda, \blambda', \btheta) = \mathcal{L}_{\text{main}}(\btheta) + \mathcal{L}_{\text{ssl}}(\blambda, \btheta) + \mathcal{L}_{\text{sup}}(\blambda', \btheta)
\end{equation}
%
%
In all unlabeled detection setups, we define $\mathcal{L}_{\text{total.}} :=  \mathcal{L}_{\text{ssl}}$ and discard the main classification task. 
%
In the rest of the paper, for the ease of notation, we use $\blambda$ to refer to all the coefficients $\{\blambda, \blambda'\}$, and we drop $\blambda, \btheta$ when it is clear from the context.  

Instead of training a separate network $f^{(n)}_\btheta$ or $f'^{(n)}_\btheta$ for each task, all the auxiliary tasks and the main classification task share a feature extraction network and each only introduces an additional 2-layer fully-connected head for each task. Training is done in a multi-task fashion in which the network is simultaneously trained for the main classification (if applicable) and all weighted auxiliary tasks using standard cross-entropy loss.

\subsection{Learning to Select Optimal Transformations} 
\label{sec:method-opt}
Previous work on self-supervised learning used ad-hoc heuristics for choosing data transformations for the training set \cite{Hendrycks2019sv,golan2018deep,tack2020csi}.
However, the optimal choice of effective transformations depends on the source distribution and heuristic approaches cannot scale up to diverse training distributions when there are many potential transformations. 
To illustrates this, we train a ResNet-18 \cite{he2016deep} with one or two self-supervised transformations that are selected from a pool of seven transformations. Here, we use the training objective in Eq. \ref{eq:eq_ssl} with equal weights for all transformations.
The OOD detection results are reported in Figure \ref{fig:all-trans}-top with CIFAR-10, CIFAR-100, and ImageNet-30 \cite{Hendrycks2019sv} datasets as in-distribution and CIFAR-100, CIFAR-10, and Pets \cite{parkhi2012cats} as example OOD test sets, respectively. The heatmap visualization presents a clear view of how different transformations (and the combinations of two) have a different impact on the OOD detection performance depending on the source distribution. 
For example, although rotation is the most effective transformation on CIFAR-10 and ImageNet-30, it is among the least effective ones for CIFAR-100. On the other hand, sharpening and color jittering are among the most effective transformations for CIFAR-100, but they perform worse on CIFAR-10.

To tackle the problem of selecting optimal transformations, we propose a simple two-step transformation selection framework. Our approach relies on Bayesian optimization to first select effective transformation set $\mathcal{T}$. It then uses meta-learning to learn $\blambda$ for OOD detection as discussed next.

\paragraph{Optimizing Transformations Set $\mathcal{T}$:} We use Bayesian optimization to identify effective transformations for each in-domain training set as the first step shown in Alg.~\ref{alg:only_alg}. Here, we assume that transformation weights $\blambda$ are equal to one and we only search for effective transformations set from a pool of available transformations. Due to the small $\mathcal{T}$ search space (i.e., $2^n$ for $n$ transformations), we use a low-cost off-the-shelf Bayesian optimization library \cite{akiba2019optuna} with Tree-Parzen estimators to find the optimum self-supervised task set. 
The Bayesian optimization objective seeks to minimize the main classification loss $\mathcal{L}_{\text{main}}$ on $D_{val}^{in}$, the validation set for the in-domain training data.

\paragraph{Optimizing Transformations Weights $\blambda$:} Next, we optimize $\blambda$ coefficients for the selected transformation from the previous step to improve the effect of shifting transformation on representation learning. 
This step is important because the $\blambda$ coefficients modulate the impact of different transformations in the training objective in Eq.~\ref{eq:eq_ssl} and Eq.~\ref{eq:eq_sup}. 
Here, we assume that $\blambda$ is a ``meta-parameter'' and we use a differentiable hyperparameter optimization algorithm~\cite{maclaurin2015gradient} for optimizing it as the second step shown in Alg.~\ref{alg:only_alg}. 
Our optimization algorithm consists of inner training updates that trains network parameters $\btheta$ using $\mathcal{L}_{\text{total}}$ on $D_{train}^{in}$ for $K$ steps. 
Given the current state of parameters $\btheta$, we update $\blambda$ in the outer loop such that $\mathcal{L}_{\text{main}}(\btheta)$ is minimized on $D_{val}^{in}$. 
Note that the gradient of $\mathcal{L}_{\text{main}}(\btheta)$ w.r.t. $\blambda$ is defined only through the gradient updates in the inner loop. Thus, the $\blambda$ updates in the outer loop require backpropagating through the gradients updates in the inner loop which can be done easily using differentiable optimizers \cite{grefenstette2019generalized}.
We use $K=1$ step for the inner-loop optimization with SGD when updating $\btheta$ and we use Adam \cite{kingma2014adam} to update $\blambda$ with small $\beta$ learning rate of 0.01. 
Figure \ref{fig:all-trans}-bottom presents $\blambda$ values during optimization from a study on three training sets.

\input{algs/main-alg}
Because the choice of effective shifting transformations depends on the in-domain distribution, our optimization framework avoids the need for $D_{test}^{out}$ samples and only relies on in-domain validation loss as a proxy for representation learning. 
Our ablation studies show that multi-task training of shifting transformations with this objective function is an effective proxy for selecting optimal transformations for both OOD detection and in-domain generalization. 

\input{tables/multiclass}

\subsection{OOD Detection Scores} \label{sec:scores}
In multi-class detection, we consider two ways for computing the detection score: (i) since all \textit{supervised heads} are trained on the same task, we get the $\lambda$ weighted sum of the softmax predictions from the main task and all auxiliary supervised transformation heads to compute an \textit{ensemble score}. (ii) Alternatively, to reduce the test-time computational complexity, a faster detection score can be computed using only the main classification head. 
Given softmax scores obtained from either (i) or (ii), in all experiments we use KL-divergence loss between the softmax scores and uniform distribution as the OOD detection score. 

In unlabeled and one-class detection with only self-supervised heads, we first get the KL-divergence between each auxiliary head and its self-labeled targets as done by \citet{Hendrycks2019sv}, then we calculate the final ensemble score using $\lambda$ weighted sum of these scores from all auxiliary heads.

%
%
%

%% file: figures/tsne-vis.tex
\begin{figure}[!t]
\begin{center}
    \begin{subfigure}[b]{0.49\columnwidth}
        \centerline{\includegraphics[width=0.99\columnwidth]{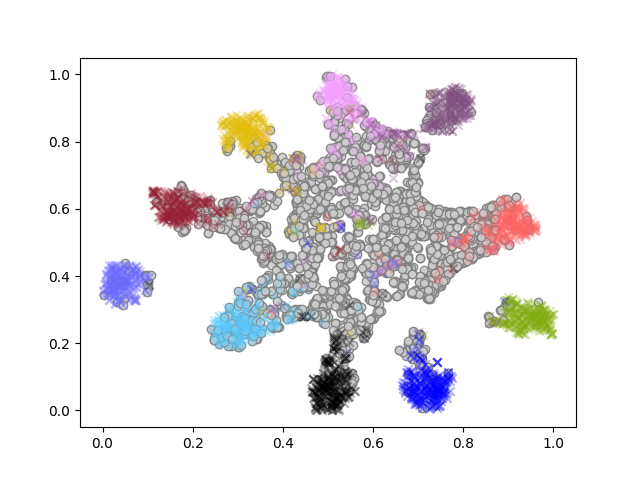}}
        \captionsetup{justification=centering} 
        \caption{Supervised Learning} 
    \end{subfigure}
    \begin{subfigure}[b]{0.49\columnwidth}
        \centerline{\includegraphics[width=0.99\columnwidth]{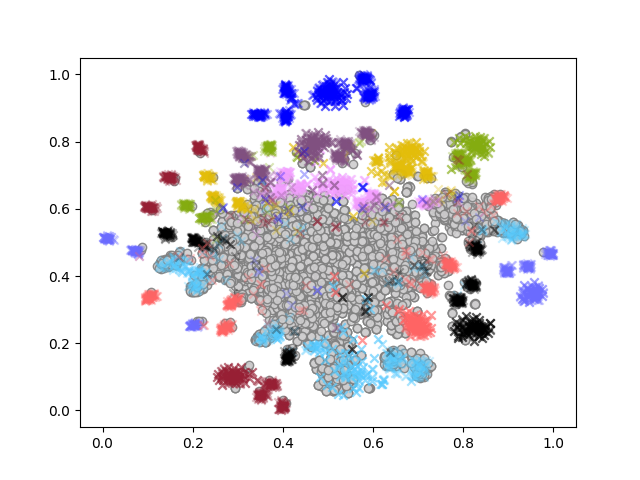}}
        \captionsetup{justification=centering}
        \caption{Shifting Transformations}
    \end{subfigure}
\caption{The t-SNE visualization of the penultimate layer features in a ResNet-18 network trained on CIFAR-10 using (a) supervised learning with cross-entropy loss and (b) our method with shifting transformation learning. OOD samples (Place365 dataset) are presented in gray.} 
\label{fig:tsne}
\end{center}
\vskip -0.2in
\end{figure}

%% file: figures/trans-set-and-weights.tex
\begin{figure*}[t]
\vskip 0.1in
\begin{center}
    \begin{subfigure}[b]{0.32\textwidth}
        \centerline{\includegraphics[width=0.97\textwidth]{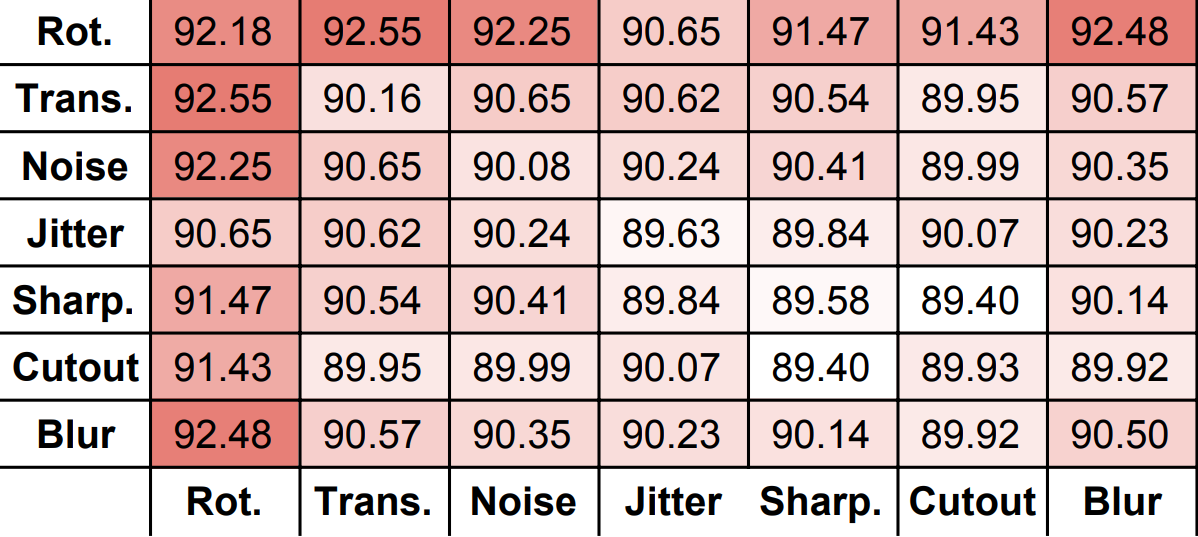}}
        \vskip 0.1in
        \centerline{\includegraphics[width=0.97\textwidth]{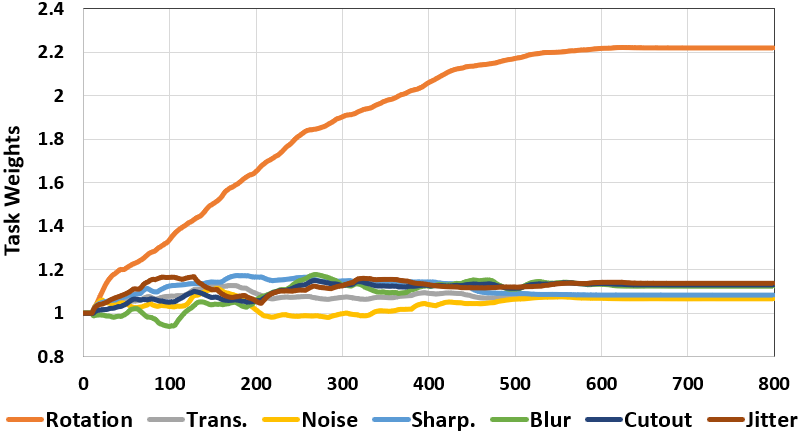}}
        \caption{CIFAR-10}
    \end{subfigure}
    \rulesep
    \begin{subfigure}[b]{0.32\textwidth}
        \centerline{\includegraphics[width=0.97\textwidth]{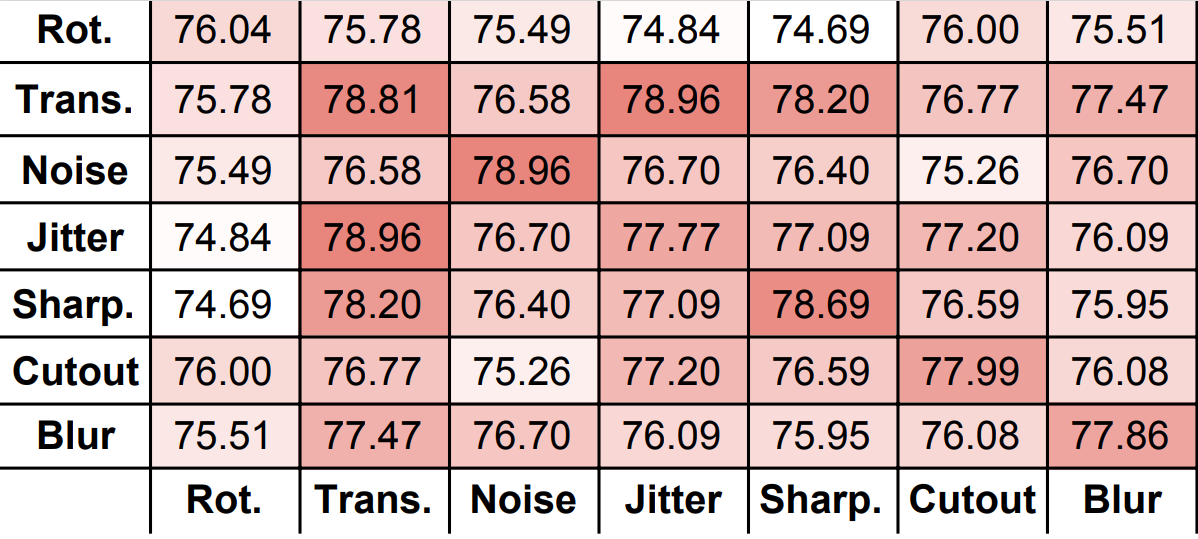}}
        \vskip 0.1in
        \centerline{\includegraphics[width=0.97\textwidth]{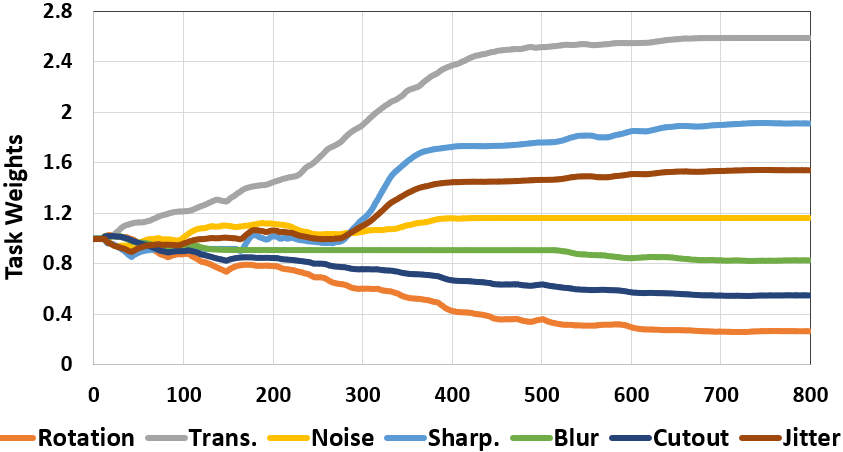}}
        \caption{CIFAR-100}
    \end{subfigure}
    \rulesep
    \begin{subfigure}[b]{0.32\textwidth}
        \centerline{\includegraphics[width=0.97\textwidth]{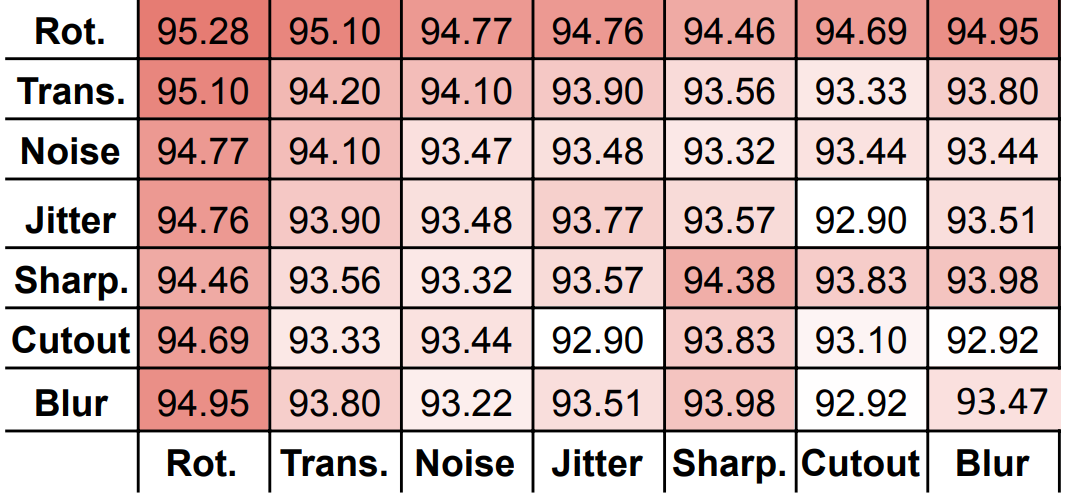}}
        \vskip 0.05in
        \centerline{\includegraphics[width=0.97\textwidth]{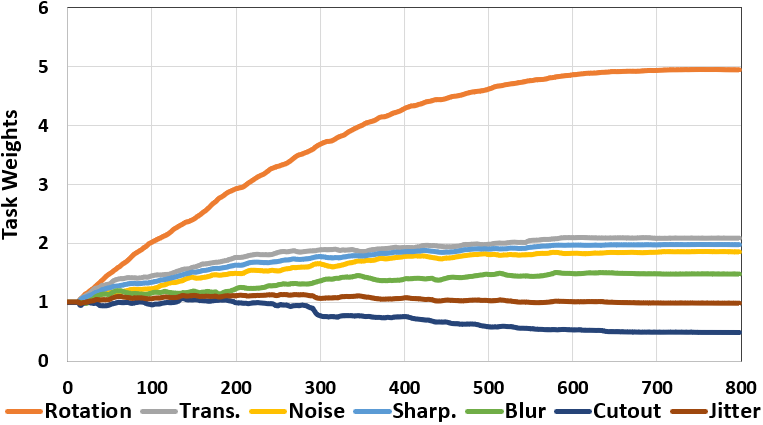}}
        \caption{ImageNet-30}
    \end{subfigure}
\caption{Our studies show that the optimal transformation set $\mathcal{T}$ and their weights $\blambda$ depend on the in-domain training set. \textbf{Top:} Ablation study to measure effects of individual and paired transformations on OOD detection performance. \textbf{Bottom:} Optimizing transformation weights ($\blambda$) for auxiliary self-supervised tasks for each training set. Experiments are done in multi-class classification setup on different training sets.}
\label{fig:all-trans}
\end{center}
\vskip -0.2in
\end{figure*}

%% file: algs/main-alg.tex
\begin{algorithm}[tb]
\caption{Transformations $\mathcal{\mathcal{T}}$ and $\lambda$ Optimization}
\label{alg:only_alg}
\textbf{Input:} Available transformation set $\mathcal{T}$, learning rate $\alpha,\beta$, inner steps $K$  \\
\textbf{Output:} Optimal $\mathcal{T}_{opt}$ and $\blambda_{opt}$ sets \\
\textbf{Step 1:} Transformations Selection 
\begin{algorithmic}[1] 
    \WHILE{not converged}
        \STATE Sample a new $\mathcal{T}$ set with $\blambda=1$.
        \STATE Train a classifier with $\mathcal{L}_{\text{total}}$ loss.
        \STATE Calculate $\mathcal{L}_{\text{main}}$ on $D_{val}^{in}$ as fitness measure.
        \STATE Update the acquisition function.
    \ENDWHILE
\end{algorithmic}
\textbf{Step 2:} $\blambda$ Weights Optimization
\begin{algorithmic}[1] 
\STATE Initialize with $\blambda = 1$.
\WHILE{not converged}
    \FOR{$K$ steps}
        \STATE  $\btheta = \btheta - \alpha\nabla_{\btheta} \mathcal{L}_{\text{total}}(\blambda, \btheta)$ on $D_{train}^{in}$
    \ENDFOR
    \STATE $\blambda = \blambda - \beta\nabla_{\blambda} \mathcal{L}_{\text{main}}(\btheta)$ on $D_{val}^{in}$
\ENDWHILE
\end{algorithmic}
\end{algorithm}

%% file: tables/multiclass.tex

\begin{table*}[t]
\centering
\caption{Comparison of OOD detection results (AUROC \%) with the supervised Baseline, state-of-the-art self-supervised \cite{Hendrycks2019sv}, contrastive learning \cite{khosla2020supervised,sehwag2021ssd,tack2020csi} and our technique with multi-task self-supervised ($\mathcal{L}_{\text{ssl}}$) and hybrid ($\mathcal{L}_{\text{ssl}}+\mathcal{L}_{\text{sup}}$) transformation learning tasks.} 
\label{tab:main-results}
\resizebox{0.82\textwidth}{!}{
\begin{tabular}{ccccclccc}
\toprule
\multirow{2}{*}{$D_{train}^{in}$} & \multirow{2}{*}{$D_{test}^{out}$} & \multicolumn{7}{c}{Detection AUROC} \\ \cline{3-9} 
 &  & Baseline & 
 \begin{tabular}[c]{@{}c@{}}Geometric\end{tabular} & 
 \begin{tabular}[c]{@{}c@{}}SupSimCLR\end{tabular} & 
 \begin{tabular}[c]{@{}c@{}}SSD+\end{tabular} & 
 \begin{tabular}[c]{@{}c@{}}CSI (ens)\end{tabular} & 
 \multicolumn{2}{c}{\begin{tabular}[c]{@{}c@{}}Ours\\($\mathcal{L}_{\text{ssl}}$) \space\space\space ($\mathcal{L}_{\text{ssl}}+\mathcal{L}_{\text{sup}}$)\end{tabular} } \\ \hline
\multicolumn{1}{c|}{\multirow{7}{*}{\rotatebox{90}{CIFAR-10}}} & \multicolumn{1}{c|}{SVHN} & 92.89 & 97.96 & 97.22 & 93.80 & 96.11 (97.38) & \textbf{99.92} & 96.60 \\
\multicolumn{1}{c|}{} & \multicolumn{1}{c|}{Texture} & 87.69 & 96.25 & 94.21 & 94.05 & 95.92 (97.18) & \textbf{97.61} & 96.91 \\
\multicolumn{1}{c|}{} & \multicolumn{1}{c|}{Places365} & 88.34 & 92.57 & 91.11 & 91.77 & 92.21 (93.11) & 93.72 & \textbf{98.73} \\
\multicolumn{1}{c|}{} & \multicolumn{1}{c|}{TinyImageNet} & 87.44 & 92.06 & 92.10 & 90.28 & 91.33 (92.49) & 92.99 & \textbf{93.57} \\
\multicolumn{1}{c|}{} & \multicolumn{1}{c|}{LSUN} & 89.87 & 93.57 & 92.13 & 94.40 & 92.91 (94.02) & \textbf{95.03} & 94.12 \\
\multicolumn{1}{c|}{} & \multicolumn{1}{c|}{CIFAR-100} & 87.62 & 91.91 & 88.36 & 90.40 & 90.60 (92.06) & 93.24 & \textbf{94.07} \\ \cline{2-9} 
\multicolumn{1}{c|}{} & \multicolumn{1}{c|}{Average} & 88.98 & 94.05 & 92.52 & 92.45 & 93.18 (94.37) & 95.42 & \textbf{95.67} \\ \hline  \hline
\multicolumn{1}{c|}{\multirow{7}{*}{\rotatebox{90}{CIFAR-100}}} & \multicolumn{1}{c|}{SVHN} & 79.18 & 83.62 & 81.55 & 83.60 & 79.22 (87.38) & 87.11 & \textbf{90.64} \\
\multicolumn{1}{c|}{} & \multicolumn{1}{c|}{Texture} & 75.28 & 82.39 & 76.83 & 81.35 & 78.33 (78.31) & \textbf{85.47} & 77.99 \\
\multicolumn{1}{c|}{} & \multicolumn{1}{c|}{Places365} & 76.07 & 74.57 & 75.37 & 79.16 & 77.15 (78.1) & 77.87 & \textbf{92.62} \\
\multicolumn{1}{c|}{} & \multicolumn{1}{c|}{TinyImageNet} & 78.53 & 77.56 & 80.77 & 76.29 & 80.07 \textbf{(82.41)} & 80.66 & 79.25 \\
\multicolumn{1}{c|}{} & \multicolumn{1}{c|}{LSUN} & 73.73 & 71.86 & 73.50 & 63.77 & 74.89 \textbf{(75.22)} & 74.32 & 74.01 \\
\multicolumn{1}{c|}{} & \multicolumn{1}{c|}{CIFAR-10} & 78.26 & 74.73 & 73.28 & 73.94 & 75.98 (78.44) & 79.25 & \textbf{91.56} \\ \cline{2-9} 
\multicolumn{1}{c|}{} & \multicolumn{1}{c|}{Average} & 76.84 & 77.46 & 76.88 & \multicolumn{1}{c}{76.35} & 77.61 (79.98) & 80.78 & \textbf{84.35} \\ \hline \hline
\multicolumn{1}{c|}{\multirow{7}{*}{\rotatebox{90}{ImageNet-30}}} & \multicolumn{1}{c|}{Flowers 101} & 87.70 & 92.13 & 93.81 & 96.47 & 95.43 (96.18) & 94.19 & \textbf{97.18} \\
\multicolumn{1}{c|}{} & \multicolumn{1}{c|}{CUB-200} & 85.26 & 90.58 & 89.19 & \textbf{96.57} & 93.32 (94.15) & 93.34 & 96.44 \\
\multicolumn{1}{c|}{} & \multicolumn{1}{c|}{Dogs} & 90.30 & 93.25 & 95.16 & 95.23 & 96.43 \textbf{(97.64)} & 93.63 & 97.07 \\
\multicolumn{1}{c|}{} & \multicolumn{1}{c|}{Food} & 78.93 & 85.09 & 83.61 & 85.48 & 88.48 (89.04) & 82.51 & \textbf{96.49} \\
\multicolumn{1}{c|}{} & \multicolumn{1}{c|}{Pets} & 92.88 & 95.28 & 96.38 & 96.24 & 97.35 \textbf{(98.49)} & 94.82 & 96.37 \\
\multicolumn{1}{c|}{} & \multicolumn{1}{c|}{Texture} & 86.98 & 92.16 & 98.70 & 94.86 & 97.63 \textbf{(98.54)} & 93.99 & 96.56 \\ \cline{2-9} 
\multicolumn{1}{c|}{} & \multicolumn{1}{c|}{Average} & 87.01 & 91.42 & 92.81 & 94.14 & 94.77 (95.67) & 92.08 & \textbf{96.69} \\ \bottomrule
\end{tabular}%
}
\vspace{-0.5em}
\end{table*}

%% file: sections/5.1.performance.tex
\vspace{-0.5em}
\section{Experiments and Results}
\label{sec:results}

We run our main experiments on ResNet-18 \cite{he2016deep} network to have a fair comparison with state-of-the-arts. 
We used seven different self-labeled transformations including rotation, translation, Gaussian noise, Gaussian blur, cutout, sharpening, and color distortion with details explained in Appendix \ref{appen:experiment}. In all experiments, both transformations set $\mathcal{T}$ and their training weights $\blambda$ are optimized using the proposed framework with the final ($\mathcal{T}$, $\blambda$) sets presented in Appendix \ref{appen:experiment}.
Unless mentioned otherwise, our main evaluation results are based on the ensembled score from available auxiliary heads.

\vspace{-0.5em}
\subsection{Comparison to State-of-the-Arts}
\label{sec:detection-perf}

\subsubsection{Multi-class Classification}
\label{sec:multi-class}

Table \ref{tab:main-results} presents our main evaluation results for multi-class classification training with Eq. \ref{eq:eq_total} on CIFAR-10, CIFAR-100, and ImageNet-30 \cite{Hendrycks2019sv} datasets each with six disjoint $D_{test}^{out}$ sets with details provided in Appendix \ref{appen:experiment}.
We compare our technique with the full supervised Baseline \cite{hendrycks2016baseline} and current state-of-the-art methods including self-supervised learning (Geometric) \cite{Hendrycks2019sv}, supervised contrastive learning (SupSimCLR) \cite{khosla2020supervised} and SSD \cite{sehwag2021ssd}, and contrasting shifted instances (CSI) \cite{tack2020csi} and with its ensembled version (CSI-ens). 
All techniques are trained on ResNet-18 network with equal training budget, and all except SSD+ use their softmax prediction as OOD detection score in multi-class classification. 
We compared the impact of both $\mathcal{L}_{\text{ssl}}$ and $\mathcal{L}_{\text{sup}} + \mathcal{L}_{\text{ssl}}$ training loss functions on OOD performance (in addition to $\mathcal{L}_{\text{main}}$ for the main classification task which is used by all the techniques in Table \ref{tab:main-results}). 
Results show our approach outperforms previous works with a large margin with both $\mathcal{L}_{\text{sup}}$ and $\mathcal{L}_{\text{ssl}}$ training objectives.
The averaged standard deviation for detection AUROC over six test sets from 5 runs of our techniques shows 0.13\% for CIFAR-10, 0.33\% for CIFAR-100, and 0.18\% for ImageNet-30.

Moreover, Table \ref{tab:main-results} shows that training on the optimized transformation set $\mathcal{T}$ and $\blambda$ weights only using an in-domain validation set consistently outperforms the previous work when testing on diverse $D_{test}^{out}$ sets. 
This observation highlights the dependency of the optimal set of shifting transformations on the in-domain training set as opposed to prior work that manually selected the shifting transformation. 
In fact, we observe that all prior work based on rotation transformation perform worse than the suervised Baseline on the CIFAR-100 experiment when testing with CIFAR-10 as the $D_{test}^{out}$ with the exception of CSI-ens.

\subsubsection{Unlabeled Detection} 
Next, we test our technique for multi-class unlabeled and one-class OOD detection trained with the $\mathcal{L}_{\text{ssl}}$ loss (Eq. \ref{eq:eq_ssl}) using our proposed transformation optimization framework. 
Table \ref{tab:one-class}(a) presents results for unlabeled multi-class detection in which averaged detection AUROC over the six $D_{test}^{out}$ sets is outperforming state-of-the-art methods with a large margin except for the CIFAR-10 experiment. 
Table \ref{tab:one-class}(b) shows detailed one-class classification results for each of the CIFAR-10 classes as $D^{in}_{train}$ and the remaining classes as $D^{out}_{test}$. 
Our technique with 90.9\% averaged AUROC on CIFAR-10 one-class detection outperforms previous works including DROCC \cite{goyal2020drocc}, GOAD \cite{bergman2020classification}, Geometric, and SSD, with the exception of CSI which requires a far more computationally expensive distance-based detection score.

\subsection{Ablation Studies}
\label{sec:ablation}
In this section, we provide additional ablation experiments to quantify contributions made by different parts of our model.

\paragraph{Transformation Optimization: } 
Table \ref{tab:tiny-opt}-top presents an analysis of the effects of our transformation set $\mathcal{T}$ and $\blambda$ weights optimizations on OOD detection when training on CIFAR-10 and testing on CIFAR-100 as $D_{test}^{out}$. 
When training with all available transformations with equal $\mathbf{\blambda = 1}$ weights (first row), the detection AUROC drops 2.76\% compared to training with both $\mathcal{T}$ and $\blambda$ optimized (forth row). 
We also observe that when only enabling $\blambda$ or $\mathcal{T}$ optimizations (second and third rows, respectively), the OOD detection performance is sub-optimal. 
We hypothesize that the gradient-based $\blambda$ optimization only has access to training signal from a few gradient updates on $\theta$ and it does not capture the full effect of mini-batch transformations on OOD detection for long training. In contrast, Bayesian optimization in $\mathcal{T}$ optimization can capture the effect of each transformation on the full training and improves OOD detection by a large margin. Nevertheless, we obtain the best OOD detection result (fourth row) with both $\blambda$ and $\mathcal{T}$ optimizations.

Our method avoids making any assumption on the availability of any $D^{out}$ for model training or hyperparameter optimization unlike prior work such as ODIN \cite{lee2017training}, Mahalanobis \cite{lee2018simple}, Outlier Exposure \cite{hendrycks2018deep} detectors. 
In the absence of OOD samples for training, we utilize the in-domain validation loss as a proxy and we examine the generalization capability of our model to unseen examples to guide the optimization of shifting transformations in Alg.~\ref{alg:only_alg}. 
In an ablation experiment in Table~\ref{tab:tiny-opt}-bottom, we use a disjoint subset of 80 Million Tiny Images dataset \cite{torralba200880} as the source of unlabeled $D^{out}$ and the KL divergence between $D^{out}$ predictions and uniform distribution as the optimization objective for both $\mathcal{T}$ and $\blambda$ optimizations. 
Interestingly, we only observe a slight performance improvement when assuming access to an auxiliary $D^{out}$ for transformation optimization.

\input{tables/one-class}

\paragraph{Advantage of Ensemble Detection Score: } 
Table \ref{tab:tiny-ens} presents OOD detection performance when using only clean samples for the main classification head compared to using transformed samples for generating an ensemble of detection scores from all auxiliary heads.
Results are based on averaged OOD detection AUROC over six $D_{test}^{out}$ sets.
We observe a clear performance increase when using the ensemble of auxiliary heads compared to the main classification head. 

\paragraph{Comparison to Data Augmentation Techniques: } A natural question to ask is whether the improvements in OOD detection could also be obtained with the supervised baseline that is trained with data augmentations. In Table \ref{tab:tiny-aug}, we compare our method with the supervised baseline trained with RandAugment \cite{cubuk2020randaugment} and AutoAugment \cite{cubuk2019autoaugment} techniques as two popular augmentation techniques. We observe that both data augmentation techniques achieve competitive OOD detection performance on ResNet-18, improving upon the supervised baseline. However, they perform significantly lower than our proposed shifting transformation learning.
Note that we could not use our shifting transformations as the data augmentation in supervised training as our transformations heavily change the input distribution and do not let training converge.


\input{tables/tiny_opt}

\input{tables/tiny_ens}

\input{tables/tiny_aug}

%% file: tables/one-class.tex
\begin{table}[]
\centering
\caption{Comparison of OOD detection results (AUROC \%) with different one-class classification and unlabeled multiclass datasets on CIFAR-10, CIFAR-100, and ImageNet-30.}
\label{tab:one-class}
(a) Unlabeled CIFAR-10, CIFAR-100, and ImageNet-30
\vskip 0.03in
\resizebox{\columnwidth}{!}{
\begin{tabular}{cccccc}
\toprule
$D^{in}$ & 
\begin{tabular}[c]{@{}c@{}}Geometric\end{tabular} & 
\begin{tabular}[c]{@{}c@{}}SimCLR\end{tabular} &  
\begin{tabular}[c]{@{}c@{}}SSD\end{tabular} &  
\begin{tabular}[c]{@{}c@{}}CSI-ens\end{tabular} &   
\begin{tabular}[c]{@{}c@{}}Ours\\($\mathcal{L}_{\text{ssl}}$)\end{tabular} \\ \toprule
\multicolumn{1}{c|}{CIFAR-10} & 86.04 & 77.84 & 84.54 & \textbf{91.99} & 89.8 \\
\multicolumn{1}{c|}{CIFAR-100} & 75.28 & 48.81 & 66.41 & 71.91 & \textbf{83.95} \\
\multicolumn{1}{c|}{ImageNet-30} & 85.11 & 65.27 & 87.42 & 92.13 & \textbf{96.57} \\
\bottomrule
\end{tabular}
}
\vskip 0.07in
(b) One-class Detection on CIFAR-10 
\vskip 0.03in
\resizebox{\columnwidth}{!}{
\begin{tabular}{ccccclcc}
\toprule
$D^{in}$ & 
\begin{tabular}[c]{@{}c@{}}DROCC\end{tabular} &  
\begin{tabular}[c]{@{}c@{}}GOAD\end{tabular} &  
\begin{tabular}[c]{@{}c@{}}Geom.\end{tabular} &  
\begin{tabular}[c]{@{}c@{}}SSD\end{tabular} &  
\begin{tabular}[c]{@{}c@{}}CSI\\-ens\end{tabular} &  
\begin{tabular}[c]{@{}c@{}}Ours\\($\mathcal{L}_{\text{ssl}}$)\end{tabular} \\ \toprule
\multicolumn{1}{c|}{Airplane} & 81.7 & 75.5 & 80.2 & 82.7 & 90.0 & 84.3 \\
\multicolumn{1}{c|}{Automobile} & 76.7 & 94.2 & 96.6 & 98.5 & 99.1 & 96.0 \\
\multicolumn{1}{c|}{Bird}& 66.6 & 82.4 & 85.9 & 84.2 & 93.3 & 87.7 \\
\multicolumn{1}{c|}{Cat}  & 67.2 & 72.1 & 81.7 & 84.5 & 86.4 & 82.3 \\
\multicolumn{1}{c|}{Deer} & 73.6 & 83.7 & 91.6 & 84.8 & 94.8 & 91.0 \\
\multicolumn{1}{c|}{Dog}  & 74.4 & 84.8 & 89.8 & 90.9 & 94.4 & 91.5 \\
\multicolumn{1}{c|}{Frog} & 74.4 & 82.8 & 90.2 & 91.7 & 94.4 & 91.1 \\
\multicolumn{1}{c|}{Horse}  & 71.3 & 93.4 & 96.1 & 95.2 & 95.2 & 96.3 \\
\multicolumn{1}{c|}{Ship}  & 80.0 & 92.6 & 95.1 & 92.9 & 98.2 & 96.3 \\
\multicolumn{1}{c|}{Truck}  & 76.2 & 85.1 & 92.8 & 94.4 & 97.9 & 92.3 \\ \hline
\multicolumn{1}{c|}{Average} & 74.2 & 84.7 & 90.0 & 90.0 & \textbf{94.3} & 90.9 \\ \bottomrule
\end{tabular}
}
\vskip -0.17in
\end{table}

%% file: tables/tiny_opt.tex

\begin{table}[]
\centering
\caption{Ablation study on $\mathcal{T}$ and $\lambda_n$ optimizations and optimization loss}
\vskip -0.05in
\label{tab:tiny-opt}
\resizebox{0.35\textwidth}{!}{
\begin{tabular}{cccc}
\hline
access to $D^{out}$ &  $\lambda_n$ opt. & $\mathcal{T}$ opt. & AUROC  \\ \hline
\textbf{--} &\textbf{--} & \textbf{--} & 90.36  \\
\textbf{--} &\ding{51} & \textbf{--} & 91.72   \\    
\textbf{--} &\textbf{--} & \ding{51} & 92.90   \\
\textbf{--} &\ding{51} & \ding{51} & \bf 93.12  \\ \midrule       
\ding{51} & \ding{51} & \ding{51} & 93.24  \\\hline  
\end{tabular}
}
\end{table}


%% file: tables/tiny_ens.tex
\begin{table}[]
\centering
\caption{Detection using only the main classification head vs. ensemble of auxiliary heads.}
\label{tab:tiny-ens}
\resizebox{0.30\textwidth}{!}{
\begin{tabular}{@{}ccc@{}}
\hline
$D^{in}$ & \begin{tabular}[c]{@{}c@{}}only \\ main head\end{tabular} & \begin{tabular}[c]{@{}c@{}}ensemble \\ aux. heads\end{tabular} \\ \midrule
\multicolumn{1}{c|}{CIFAR-10} & 93.83 & \bf 95.67 \\ 
\multicolumn{1}{c|}{CIFAR-100} & 79.30 & \bf 84.35 \\ 
\multicolumn{1}{c|}{ImageNet-30} & 92.59 & \bf 96.69 \\ \midrule
\end{tabular}%
}
\vskip -0.17in
\end{table}


%% file: tables/tiny_aug.tex
\begin{table}
\centering
\vskip -0.15in
\caption{Detection AUROC with our technique vs. data augmentation.}
\label{tab:tiny-aug}
\resizebox{0.4\textwidth}{!}{
\begin{tabular}{@{}ccccc@{}}
\hline
$D^{in}$ & Baseline & AutoAug. & RandAug. & Ours \\ \midrule
\multicolumn{1}{c|}{CIFAR-10} & 88.98 & 92.46 & 92.72 & \bf 95.67 \\ 
\multicolumn{1}{c|}{CIFAR-100} & 76.84 & 78.68 & 78.62 & \bf 84.35 \\ \midrule
\end{tabular}%
}
\vskip -0.15in
\end{table}

%% file: sections/5.2.robustness.tex

\input{tables/criteria}


\vspace{-0.2em}
\section{OOD Detection Generalizability}
\label{sec:detection-robust} 

In real-world application, we characterize four main criteria required from an ideal OOD detection technique, including i) zero-shot OOD training, ii) no hyperparameter dependency, and iii) generalization to various unseen OOD distributions and iv) robustness against test-time perturbations. 
In this section, we situate our proposed technique against a diverse range of state-of-the-art OOD detection techniques along these requirements. 
Table \ref{tab:table-of-criteria} presents results for training on CIFAR-10 and testing on six $D_{test}^{out}$ sets used in Table \ref{tab:main-results}. 
Note that this is not intended to be a ranking of different OOD detection techniques; instead, we aim to review trade-offs and limitations among different detection approaches.

\paragraph{Hyperparameters Dependency: } While hyperparameter tuning for the training of in-domain samples is done using a held-out validation set, the hyperparameter disentanglement is a crucial property for OOD detection. 
Specifically, an ideal detector should not be sensitive to hyperparameters tied to the target outlier distribution.
Table \ref{tab:table-of-criteria} divides different techniques w.r.t their dependency on detection hyperparameters into three levels of high, low, and no dependency. 
Techniques with high dependency like ODIN \cite{liang2017enhancing} and Mahalanobis \cite{lee2018simple} use a validation set of $D^{out}$ for training, resulting in poor performance under unseen or diverse mixture of outlier distributions. 
Table \ref{tab:table-of-criteria} shows over 3\% performance gap between the averaged detection performance on six $D_{test}^{out}$ sets (Column 5) and detection performance under an equal mixture of the same test sets (Column 6) for these two detectors which indicates strong $D^{out}$ hyperparameter dependency. 

Techniques with low dependency do not use a subset of $D^{out}_{test}$, however, they depend on hyperparameters such as the choice of $D^{out}_{train}$ set (e.g., Outlier Exposure \cite{hendrycks2018deep}), or hand-crafted self-supervised tasks  \cite{golan2018deep}, \cite{Hendrycks2019sv}, or data augmentation  \cite{tack2020csi} that requires post training $D^{out}_{test}$ for validation. 
These techniques can suffer significantly in settings in which the new source training set is invariant to previous hand-crafted self-supervised tasks and augmentations as seen in Figure \ref{fig:all-trans}. 
On the other hand, techniques with no hyperparameters like Gram Matrices \cite{sastry2019detecting}, SSD \cite{sehwag2021ssd}, and our proposed framework bears no hyperparameter dependency on the choices of in-domain or outlier distribution. 
Note that many techniques, like ours, use a $\blambda$ training hyperparameter to balance training between in-domain classification and auxiliary tasks. However, in our case these hyperparameters are tuned automatically without requiring OOD training samples.

\paragraph{Zero-shot Training: } 
A previous trend in OOD detection techniques considered using a subset of the target $D^{out}_{test}$ for model tuning (e.g., ODIN \cite{liang2017enhancing} and Mahalanobis \cite{lee2018simple}) or using an auxiliary $D^{out}_{train}$ set as a part of model training (e.g., Outlier Exposure \cite{hendrycks2018deep}). 
Although these techniques can achieve high detection performance with the right training set, having access to the specific $D^{out}_{tune}$ for tuning or even any $D^{out}_{train}$ for training the detector is not a realistic assumption in practical setups.
An efficient proposal to use these techniques is to integrate them into zero-shot techniques as presented by \cite{sastry2019detecting, Hendrycks2019sv} when $D^{out}_{train}$ is available or to benefit from taking semi-supervised or few-shot approaches as done by \cite{ruff2019deep,sehwag2021ssd}.

\paragraph{Detection Generalizability:} Recent work on OOD detection recognized the necessity of diverse $D^{out}_{test}$ sets to evaluate the generalizability of OOD detection techniques  \cite{mohseni2020ood,winkens2020contrastive,sastry2019detecting}.
Typically, near-OOD and far-OOD sets are chosen based on the semantic and appearance similarities between the in-domain and outlier distributions and, in some cases, measured by relevant similarity metrics (e.g., confusion log probability \cite{winkens2020contrastive}).
Following the previous works, we chose CIFAR-100 as the near-OOD test distribution and SVHN as the far-OOD test distribution for detectors trained on CIFAR-10. 
While Table \ref{tab:table-of-criteria} shows high performance on far-OOD for all techniques, Gram Matrices, Mahalanobis, and ODIN show 20.5\%, 10.9\%, and 10.6\% detection performance drop for near-OOD distribution compared to the far-OOD test distribution, respectively.
In comparison, our technique shows 2.53\% performance gap between far-OOD and near-OOD test distributions.

\paragraph{Detection Robustness:} Evaluating the effects of distribution shift on predictive uncertainty have been previously studied in \cite{ovadia2019can,goyal2020drocc} for real-world application. In Appendix \ref{appen:robustness_results},
we investigate the effect of natural perturbations and corruptions proposed in \cite{hendrycks2019benchmarking} on OOD detection performance. 
We measure averaged OOD detection results for all 15 image distortions on 5 levels of intensity where both $D^{in}_{test}$ and $D^{out}_{test}$ are treated with the same distortion type and level. 
Figure \ref{fig:robustness} in Appendix \ref{appen:robustness_results} presents detailed OOD detection results in which all techniques show more performance drop at the higher levels of perturbation intensity.
However, distance-based detectors (Figure \ref{fig:robustness}-a) like Gram and Mahalanobis show significantly less performance drop (4.23\% and 5.24\% AUROC drop, respectively) compared to classification-based detectors (Figure \ref{fig:robustness}-b) like Outlier Exposure and Geometric with over 14\% AUROC drop. 
Our experiments indicate the advantage of distance-based detection methods in OOD detection under test-time input perturbations.




%% file: tables/criteria.tex
\begin{table*}[]
\centering
\caption{Review of OOD detection criteria, averaged detection performance, and generalizability to unseen OOD test distributions (AUROC \%) for a diverse set of OOD detection techniques. We compare our technique with ODIN \cite{lee2017training}, Mahalanobis \cite{lee2018simple}, Outlier Exposure \cite{hendrycks2018deep}, Geometric \cite{Hendrycks2019sv}, CSI \cite{tack2020csi}, and Gram \cite{sastry2019detecting}.}
\label{tab:table-of-criteria}
\resizebox{0.8\textwidth}{!}{%
\begin{tabular}{cccccccc}
\toprule
\multirow{2}{*}{\begin{tabular}[c]{@{}c@{}}Detection \\ Technique\end{tabular}} & \multicolumn{3}{c}{OOD Detection Criteria} & \multirow{2}{*}{\begin{tabular}[c]{@{}c@{}}Averaged \\ Detection \\ Performance\end{tabular}} & \multicolumn{3}{c}{Generalizability Tests} \\ \cline{2-4} \cline{6-8} 
 & \begin{tabular}[c]{@{}c@{}}Hyp.-Para. \\ Dependency\end{tabular} & Generalizable & \begin{tabular}[c]{@{}c@{}}Zero\\Shot\end{tabular} &  & \begin{tabular}[c]{@{}c@{}}Mixed\\ Distribution\end{tabular} & Far-OOD & Near-OOD \\ \toprule
\multicolumn{1}{c|}{ODIN} & High & \textbf{--} & \textbf{--} & 91.15 & 88.10 & 96.70 & 85.80 \\
\multicolumn{1}{c|}{Mahalanobis} & High & \textbf{--} & \textbf{--} & 95.35 & 92.24 & 99.10 & 88.51 \\
\multicolumn{1}{c|}{Outlier Exposure} & Low & \ding{51} & \textbf{--} & 96.24 & 96.88 & 98.76 & 93.41 \\ 
\multicolumn{1}{c|}{Geometric} & Low & \ding{51} & \textbf{\ding{51}} & 94.05 & 94.29 & 97.96 & 91.91 \\
\multicolumn{1}{c|}{CSI-ens} & Low & \ding{51} & \textbf{\ding{51}} & 94.37 & 94.10 & 97.38 & 92.06 \\
\multicolumn{1}{c|}{SSD} & No & \ding{51} & \textbf{\ding{51}} & 92.45 & 92.70 & 93.80 & 90.40 \\
\multicolumn{1}{c|}{Gram Matrices} & No & \textbf{--} & \textbf{\ding{51}} & 94.17 & 95.08 & 99.50 & 79.01 \\
\multicolumn{1}{c|}{Ours} & No & \ding{51} & \textbf{\ding{51}} & 95.67 & 95.55 & 96.60 & 94.07 \\
\bottomrule
\end{tabular}%
}
\vskip -0.1in
\end{table*}

%% file: sections/7-conclusion.tex
\section{Conclusion} 
\label{sec:conclusion}

Developing reliable and trustworthy machine learning algorithms for open-world and safety-critical applications poses a great challenge. In this paper, we presented a simple framework for OOD detection that leverages representation learning with shifting data transformations, and we empirically demonstrated its efficacy on several image datasets. We showed that the optimal choice of shifting transformation depends on the in-domain training distribution and we propose a framework to automatically choose the optimal transformations for a given in-domain set without requiring any OOD training samples. 
Albeit its simplicity, our proposed method outperforms the state-of-the-art OOD detection techniques and exhibits strong generalization to different outlier distributions. 
A limitation of our work is longer training time and large memory requirement due to the large training batch size. Future work is focused on improving the efficiency and scalability of shifted transformation learning for larger datasets.

%% file: sections/9-appendix.tex
\newpage


\vspace{-0.5em}
\section{Experiments Details}
\label{appen:experiment}

\paragraph{Dataset details:} Our experiments are focused on image domain and we use CIFAR-10 \cite{krizhevsky2009learning}, CIFAR-100 \cite{krizhevsky2009learning}, and ImageNet-30 \cite{Hendrycks2019sv} in multi-class and unlabeled detection. 
CIFAR-10 and CIFAR-100 consist of 50,000 training and 10,000 test samples, respectively. 
ImageNet-30 is a selection of 30 classes from ImageNet \cite{deng2009imagenet} dataset that contains 39,000 training and 3,000 test samples. 
In one-class classification, we only used single classes of CIFAR-10 as training source ($D_{train}^{in}$) and the other classes as test set  ($D_{test}^{out}$). 
All $D_{test}^{out}$ test images are resized to the $D_{train}^{in}$ image size which is 32x32 in all CIFAR-10/100 experiments. 
In ImageNet-30 experiments, we first resize the images to 256x256 and then center crop to 224x224.

\paragraph{Training Details:} All experiments are based on ResNet-18 network with mini-batch size of 64, SGD optimizer with momentum of 0.9, initial learning rate of 0.1 (decayed using a cosine annealing schedule). 
Despite the set of shifting transformations, we still use a few data invariant ``native augmentations'' including random horizontal flip and small random crop and padding for the main supervised head.  
We use cross-entropy loss for all supervised and self-supervised branches with labels generated for self-supervised tasks. 
We apply all transformation targets (e.g., all four rotations for the rotation transformation) from $\mathcal{T}$ on every mini-batch during the training, and therefore, the final mini-batch is the base mini-batch size multiplied by the total number of shifting transformations. 
We observe that learning multiple shifting transformations benefits from longer training time (similar to contrastive learning setups). So we train all multi-class classification models for 800 epochs (i.e., Table 1) and unsupervised models for 200 epochs (i.e., Table 2) . 
Ablation studies presented in Figure 2 and Table 3 are trained for 200 epochs.

\paragraph{$\mathcal{T}$ and $\blambda$ Optimization:} As described in Algorithm \ref{alg:only_alg}, we first run the Bayesian optimization to find the optimum $\mathcal{T}$ set, followed by the meta-learning optimization to optimize all training $\lambda$ weights. 
Despite efforts on solely using meta-learning optimization for finding the optimum $\mathcal{T}$ set and $\lambda$ weights, we found that gradient-based optimization is not able to capture long effect of shifting transformations on OOD performance. 
Instead, given the small search space for the small number of image transformations, Bayesian optimization served well for this task. 
Note that this is only the search step, and hence we use a short training with 50 epochs on each iteration.

Our framework finds the following transformation and weight pairs for the CIFAR-10 dataset \{(Jitter, 3.2791), (Rotation, 2.7547), (Sharpening, 2.6906)\} with $\lambda_0 = 4.0760$. In CIFAR-100 dataset: \{(blur, 4.3051), (Jitter, 2.2612), (Translate, 2.9636), (Sharpening, 3.9634)\} with $\lambda_0 = 8.6546$. In ImageNet-30: \{(Noise, 5.3806), (Rotation, 3.3754), (Sharpening, 4.7626)\} with $\lambda_0 = 9.3599$.

\paragraph{Transformations Details} In contrast to common data augmentations, we followed~\cite{dosovitskiy2014discriminative} to apply transformations to the extreme degree. 
We used 7 different self-labeled transformations including rotation 
$(\{0^{\circ}, 90^{\circ}, 180^{\circ}, 270^{\circ}\})$
, translation 
(combinations of \textpm\ \SI{30}{\percent} horizontal and vertical translations)
, Gaussian noise (with standard deviations of \{0, 0.3, 0.5, 0.8\}), Gaussian blur (with sigmas of \{0, 0.3, 0.5, 0.8\}), cutout (rectangular cut with sizes of \{0, 0.3, 0.5, 0.8\}), sharpening (image blended with its convolution-based edges with alphas of \{0, 0.3, 0.6, 1.0\}), and color distortion (jittered brightness, contrast, saturation, and hue by rates of \{0, 0.4, 0.6, 0.8\}).

\input{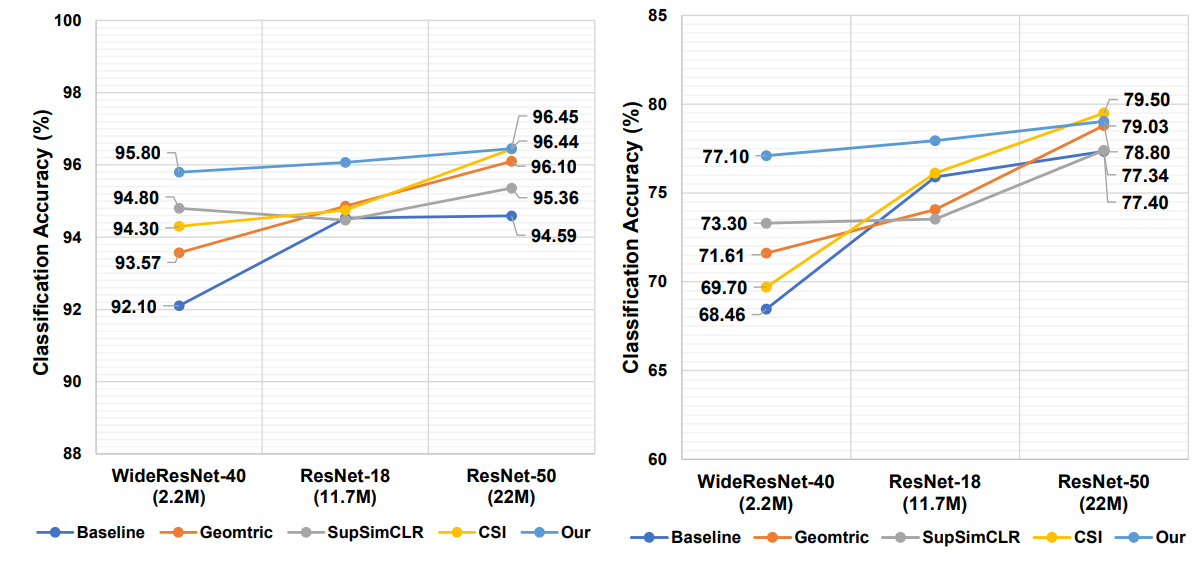} 

\paragraph{Evaluation Setup and Metrics:} We evaluate the OOD detection performance using multiple diverse $D_{test}^{out}$ sets to determine how well the detector can generalize on these unseen distributions, including test sets of SVHN \cite{netzer2011reading}, TinyImageNet ~\cite{russakovsky2015imagenet}), Places365~\cite{zhou2017places}, LSUN~\cite{yu2015lsun}, and CIFAR-10 (or CIFAR-100 when CIFAR-10 is the source training set) for CIFAR-10 and CIFAR-100 experiments and Pets \cite{parkhi2012cats}, 
Flowers-101 \cite{nilsback2006visual}), CUB-200 \cite{wah2011caltech}, Dogs \cite{khosla2011novel}, Food \cite{bossard2014food} for ImageNet-30 experiments.
We choose the area under the receiver operating characteristic curve (AUROC) \cite{davis2006relationship} as a threshold agnostic metric in all evaluations. 
The AUROC will be 100\% for the perfect detector and 50\% for a random detector. 
In all evaluations, we use $D_{test}^{out}$ (test set of the outlier dataset) as positive OOD samples and the $D_{test}^{in}$ (test set of source training dataset) as negative samples for detection.


\input{tables/perturbations}

\subsection{Classification Accuracy}
\label{sec:model-acc}
Since our technique enjoys from learning domain knowledge through multiple data transformations, we first compare the supervised classification accuracy of our technique with state-of-the-art self-supervised and contrastive learning techniques.
Figure \ref{fig:accuracy-comparison} presents classification accuracy of CIFAR-10 and CIFAR-100 datasets trained on WideResNet-40-2, ResNet-18, and ResNet-50 networks. 
Our technique outperforms other techniques across both datasets in WideResNet and ResNet-18 networks and achieves competitive performance in ResNet-50, which indicates the effectiveness of transformation learning for better generalization.  
The improvement is more visible in smaller network sizes like WideResNet-40-2 (e.g., 3.7\% gain over Baseline in CIFAR-10) and the smaller training sets (e.g., 8.64\% gain over Baseline in CIFAR-100).


\section{Additional Robustness Results}
\label{appen:robustness_results}

Figure \ref{fig:robustness} presents a robustness evaluation for different distance-based and classification-based techniques under test-time perturbations. 
In this experiment, both in-domain and OOD samples are perturbed with 15 natural perturbations on 5 levels of intensity proposed by \cite{hendrycks2019benchmarking}. 
Our results indicate that all techniques show more performance drop at the higher levels of perturbation intensity. 
Specifically, distance-based detectors show significantly less performance drop compared to classification-based detectors like Outlier Exposure and Geometric with over 14\% AUROC drop. 
Noticeably, the Mahalanobis detector with access to perturbed $D^{out}_{test}$ samples for tuning maintains fairly high detection performance under all perturbation types. 

\subsection{Computation Complexity: } We emphasize the importance of test-time computation costs in real-world applications with resource and time constraints. 
Classification-based techniques tend to perform faster as they only use the model prediction probability from the main classification or an ensemble of multiple tasks.  
For example, the Baseline, Outlier Exposure, and ODIN technique only use the model softmax prediction as the OOD detection score.
However, distance-based methods carry an overhead to measure the unknown $D_{test}^{out}$ samples' distance from the seen $D_{train}^{in}$ set. 
For instance, the CSI-ens \cite{tack2020csi} technique uses an ensemble of distances (cosine similarity distance) from multiple augmented copies of the $D_{test}^{out}$ to the entire or a subset of $D_{train}^{in}$. 
On the other hand, SSD \cite{sehwag2021ssd} detector measures the Mahalanobis distance between $D_{test}^{out}$ and a pre-trained representation of $D_{train}^{in}$ based on k-mean clustering which significantly improves the detection time. 
Table \ref{tab:tiny-time} presents a comparison between detection inference time on a ResNet-18 network trained on CIFAR-10 running on the same system with a single RTX 1080ti GPU. 
Inference time is measured for the entire $D_{test}^{out}$ set based on equal mixture of five test set. 
We encourage future research to investigate opportunities for distance-based and classification-based detectors in different applications of OOD detection.

\input{tables/tiny_time}

%% file: figures/accuracy.tex
\begin{figure}[t]
\centerline{\includegraphics[width=0.8\columnwidth]{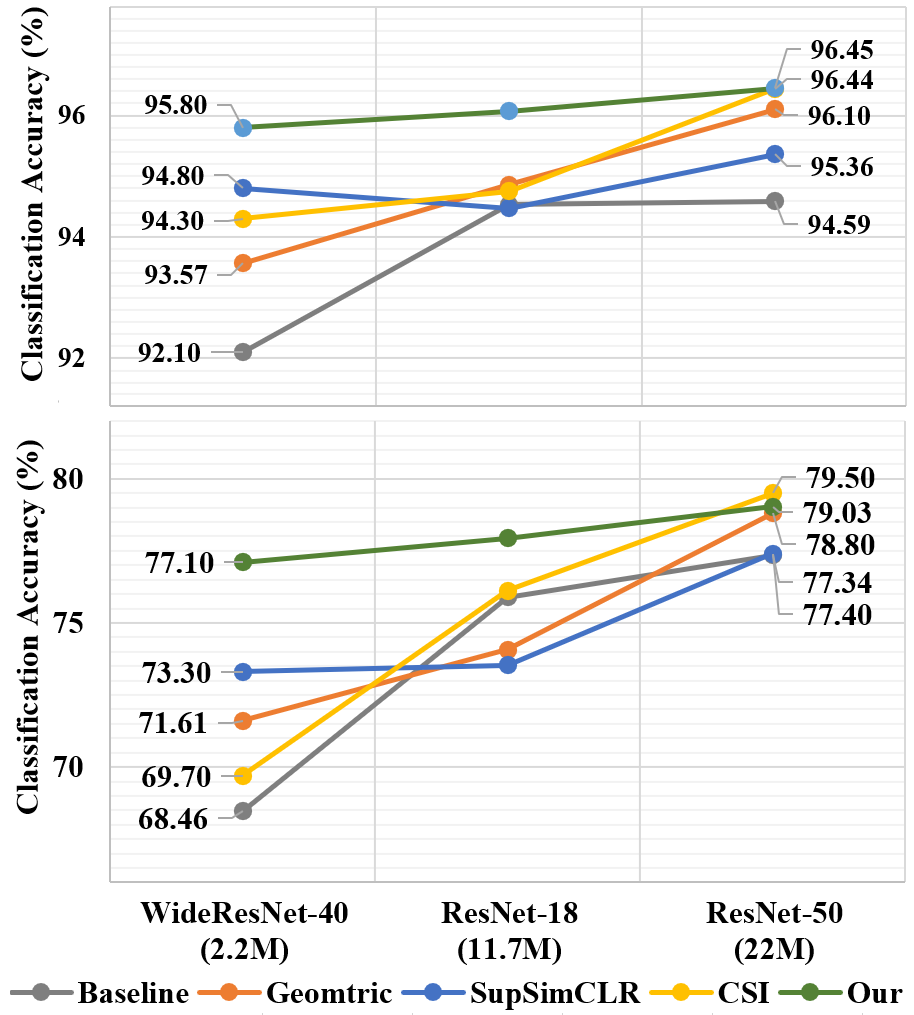}}
\caption{Classification accuracy on CIFAR-10 \textbf{(Top)} and CIFAR-100 \textbf{(Bottom)} datasets.} 
\label{fig:accuracy-comparison}
\end{figure}

%% file: tables/perturbations.tex
\begin{figure*}[t]
\begin{center}
    \centerline{\includegraphics[width=1.99\columnwidth]{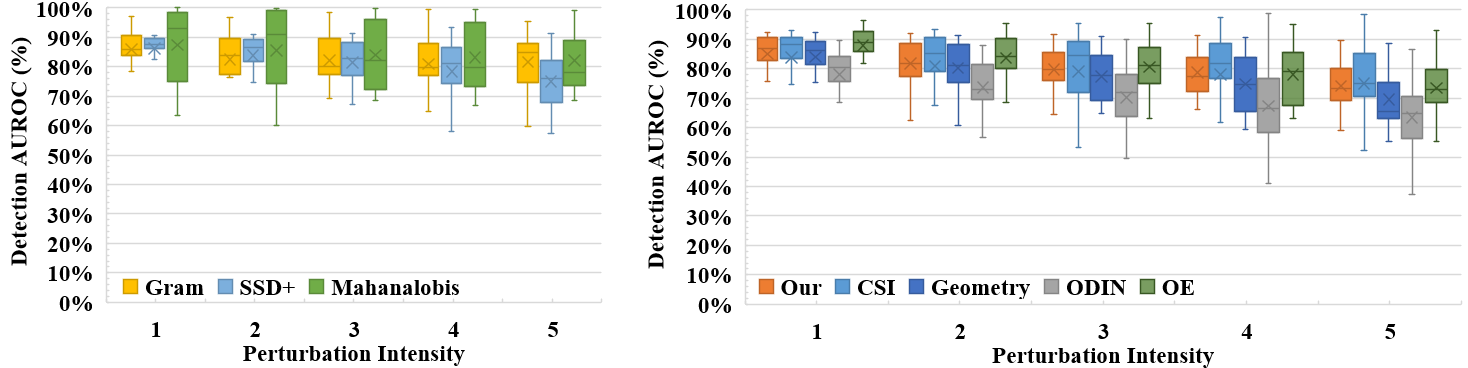}}
(a) Distance-based techniques \space\space\space\space\space\space\space\space\space\space\space\space\space\space\space\space\space\space\space\space\space
(b) Classification-based techniques
\caption{OOD detection robustness results for \textbf{(Left)} distance-based and \textbf{(Right)} Classification-based techniques with both $D_{test}^{out}$ and $D_{test}^{in}$ perturbed under 5 levels of intensity. $\times$ sign represents the mean AUROC.} 
\label{fig:robustness}
\end{center}
\end{figure*}


%% file: tables/tiny_time.tex

\begin{table}
\centering
\caption{Detection inference time averaged on five $D_{test}^{out}$ sets.}
\label{tab:tiny-time}
\resizebox{0.27\textwidth}{!}{
\begin{tabular}{@{}cc@{}}
\toprule
Detector & Inference Time (s) \\ \midrule
\multicolumn{1}{c|}{Baseline} & 9.3s \\ 
\multicolumn{1}{c|}{OE} & 9.3s \\ 
\multicolumn{1}{c|}{Geometric} & 39.1s \\ 
\multicolumn{1}{c|}{Ours} & 76.3s \\
\multicolumn{1}{c|}{SSD} & 23.6 \\ 
\multicolumn{1}{c|}{CSI} & 18.7s\\ 
\multicolumn{1}{c|}{CSI-ens} & 163.2s \\ 
\multicolumn{1}{c|}{Gram} & 323.9 \\ \midrule 
\end{tabular}%
}
\end{table}